\documentclass{article}

\PassOptionsToPackage{numbers, compress}{natbib}

\usepackage[preprint]{neurips_2025}

\usepackage[utf8]{inputenc} 
\usepackage[T1]{fontenc}    
\usepackage{hyperref}       
\usepackage{url}            
\usepackage{booktabs}       
\usepackage{amsfonts}       
\usepackage{nicefrac}       
\usepackage{microtype}      
\usepackage{xcolor}         
\usepackage{graphicx}
\usepackage{amsmath}
\usepackage{amssymb}

\usepackage{amsmath}

\hypersetup{colorlinks = true,
linkcolor = purple,
urlcolor  = blue,
citecolor = cyan,
anchorcolor = black}

\title{KinTwin: Imitation Learning with Torque and Muscle Driven Biomechanical Models Enables Precise Replication of Able-Bodied and Impaired Movement from Markerless Motion Capture}

\author{\href{https://orcid.org/0000-0001-5714-1400}{\includegraphics[scale=0.06]{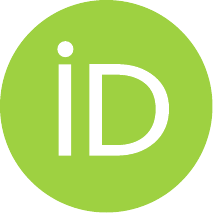}}\hspace{1mm}  R. James Cotton \\
  Department of PM\&R\\
  Northwestern University\\
  Shirley Ryan AbilityLab \\
  \texttt{rcotton@sralab.org} \\}

\begin{document}
\maketitle

\begin{abstract}
Broader access to high-quality movement analysis could greatly benefit movement science and rehabilitation, such as allowing more detailed characterization of movement impairments and responses to interventions, or even enabling early detection of new neurological conditions or fall risk. While emerging technologies are making it easier to capture kinematics with biomechanical models, or how joint angles change over time, inferring the underlying physics that give rise to these movements, including ground reaction forces, joint torques, or even muscle activations, is still challenging. Here we explore whether imitation learning applied to a biomechanical model from a large dataset of movements from able-bodied and impaired individuals can learn to compute these inverse dynamics. Although imitation learning in human pose estimation has seen great interest in recent years, our work differences in several ways: we focus on using an accurate biomechanical model instead of models adopted for computer vision, we test it on a dataset that contains participants with impaired movements, we reported detailed tracking metrics relevant for the clinical measurement of movement including joint angles and ground contact events, and finally we apply imitation learning to a muscle-driven neuromusculoskeletal model. We show that our imitation learning policy, KinTwin, can accurately replicate the kinematics of a wide range of movements, including those with assistive devices or therapist assistance, and that it can infer clinically meaningful differences in joint torques and muscle activations. Our work demonstrates the potential for using imitation learning to enable high-quality movement analysis in clinical practice.
\end{abstract}

\section{Introduction}

How people move provides a huge amount of information about their health. This is rarely quantified in clinical practice because of the technical challenges to easily measuring it. Recent advances in markerless motion capture and biomechanical modeling are lowering this barrier, allowing kinematics to be obtained through easily accessible technologies \citep{kanko_concurrent_2021, uhlrich_opencap_2023, cotton_differentiable_2025, peiffer_fusing_2024}. Kinematics, or how the joint angles and body position change over time, only tells part of the story. Obtaining kinetics, a.k.a., the torques and ground reaction forces, or even muscle activations, would be even better. For example, the propulsion ratio between the impaired and unimpaired leg of people with a hemiparetic gait after a stroke can predict responses to specific interventions, and muscle activation patterns can track responses to interventions \citep{awad_these_2020, clark_merging_2010, routson_influence_2013}.

Imitation learning from large human movement datasets is one promising solution to this challenge \citep{song_deep_2021}. It trains a goal-conditioned policy using reinforcement learning to copy movements in a physics simulator, thus solving the inverse dynamics to compute the forces the model must exert to reproduce that trajectory. Prior works in imitation learning are limited for our application in several ways: Firstly, the vast majority are based on the skinned multi-person linear (SMPL) model \citep{loper_smpl:_2015}, which has a kinematic chain that does not match biomechanical models \citep{keller_skin_2023}. Secondly, reported metrics are typically mean-per-joint-position-error, rather than biomechanical errors like joint angle tracking and foot contact timing. Lastly, they are trained and tested on datasets of movements from able-bodied individual, thus their generalization and sensitivity when applied to impaired movement have not been evaluated.

Our work addresses these limitations by training an imitation learning policy on a large dataset of 34 hours of data from 467 participants. Most have movement impairments from various etiologies, including using a lower limb prosthesis or a history of neurologic injuries. They performed a variety of movements, focusing on clinical outcome assessments for mobility. To perform imitation learning on this large dataset, we took advantage of the recent GPU acceleration of neuromechanical models in MuJoCo MJX \citep{todorov_mujoco_2012}, as well as the recent community growth developing these biomechanical models, such as MyoSuite and LocoMuJoCo  \citep{caggiano_myosuite_2022, al-hafez_locomujoco_2023}.

\begin{figure}[!htbp]
\centering
\includegraphics[width=0.8\linewidth]{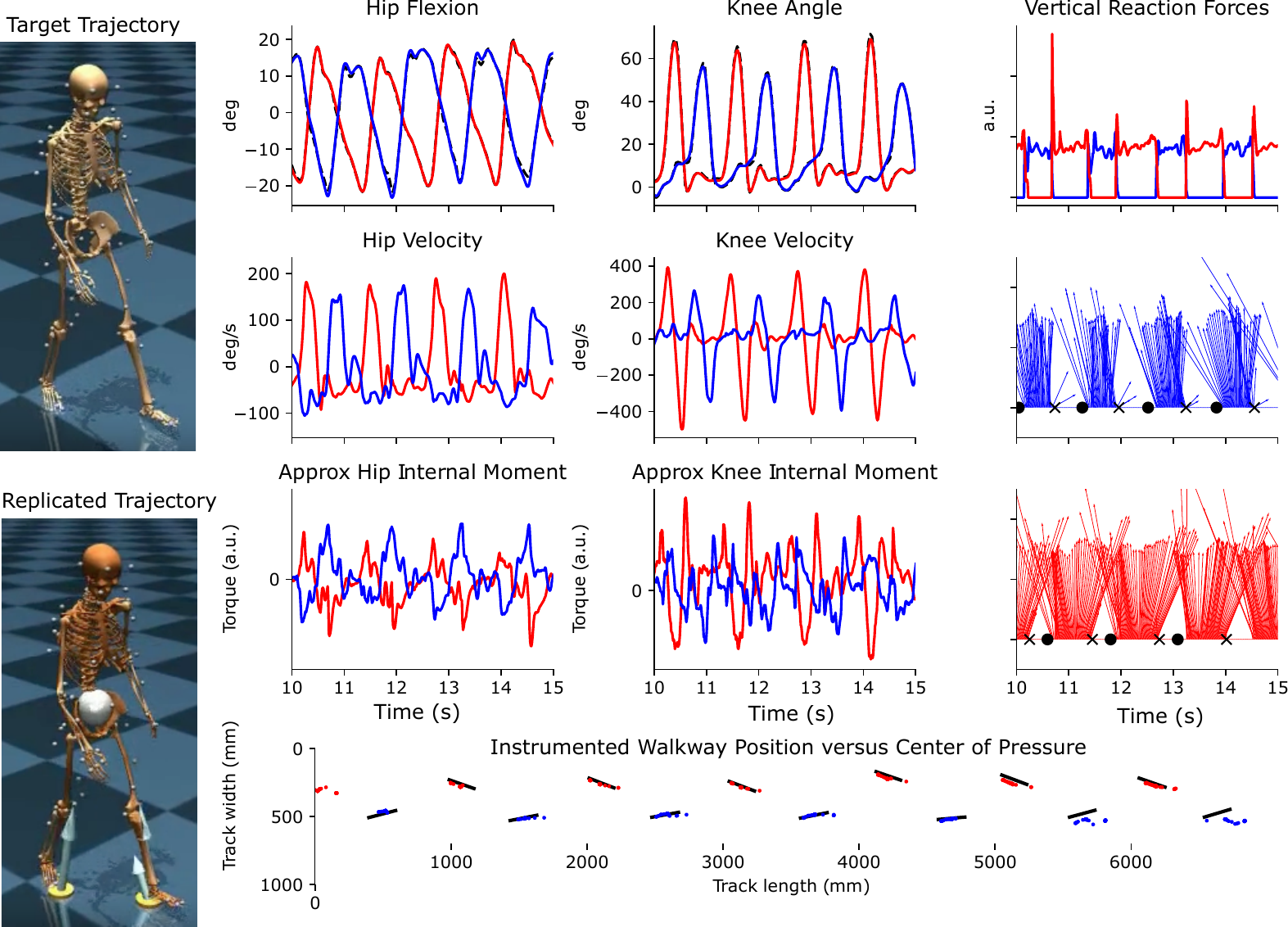}
\caption[]{Waveforms from a single trial from a participant using a left transfemoral prosthesis. Our color convention is blue for left and red for right throughout. The top row left and center columns show the target trajectory hip and knee angles respectively, in a dashed black trace, which is hardly visible due to the close tracking. The next rows show the hip and knee velocities. The last row shows the internal moments. The right column shows the ground reaction forces and detects greater forward propulsion at terminal stance from the intact compared to the prosthetic side. The black dots correspond to gait events from the instrumented walkway, and the crosses correspond to foot off events. The bottom row compares the progression of the center of pressure from the physics simulation to the line between heel and toe positions from the instrumented walkway, showing very close agreement.}
\label{fig:168_torque}
\end{figure}

Our study shows that imitation learning can be successfully applied to biomechanical models and accurately replicate the movements of people with and without movement impairments Figure~\ref{fig:168_torque}. To our knowledge, this is the first demonstration of whole-body imitation learning applied to biomechanical models. Furthermore, we demonstrate it with a torque-driven model and a neuromuscular model containing 92 muscles in the lower limbs. We validate the outputs from our policy by comparing the ground reaction forces to the gait events detected from an instrumented walkway. We also show that both the torques and muscle activations vary based on the clinical condition of participants. Our resulting system, KinTwin, produces a digital kinetic twin replicating movements captured with markerless motion capture.

In short, our contributions are:

\begin{itemize}
\item We provide the first demonstration of whole-body imitation learning applied to torque-drive and muscle-driven biomechanical models.
\item This demonstrates the power of the recently released GPU acceleration of muscle-driven musculoskeletal models in MuJoCo MJX, enabling massively accelerated and parallel neuromuscular simulations.
\item We show that this works when applied to a biomechanical dataset of movement from people with and without movement impairments. Furthermore, it can pick up clinically meaningful differences corresponding to the movement impairments.
\item We validate the ground reaction forces from our imitation learning policy against measurements from an instrument walkway and establish that it reproduces accurate spatiotemporal gait parameters during walking.
\end{itemize}

\section{Related works}

Reinforcement learning (RL) with physics-driven simulators learns policies to control models of human movement \citep{shimada_physcap_2020, shimada_neural_2021, shi_motionet_2020}.
\citet{peng_sfv:_2018, peng_deepmimic:_2018} trained a simple humanoid model to replicate human motions acquired from motion capture and even video. This approach has been extended to biomechanically accurate models controlled by simulated muscles \citep{lee_scalable_2019, song_deep_2021}. These early efforts were akin to traditional biomechanical pipelines and must be optimized for specific trajectories. More recent imitation learning policies can replicate many target trajectories.
\citep{yuan_residual_2020, yuan_simpoe_2021, luo_perpetual_2023, luo_universal_2024}. \citet{yuan_residual_2020} showed that differences between the simulated model and real biomechanics can make exact replication of motions impossible, but can be compensated by residual forces applied to the center of mass.
\citet{yuan_simpoe_2021} (SimPoE) extended this to a policy that tracks movements from monocular video. Their impressive results showed that the constraints of physics reduced physically implausible artifacts such as jerky, discontinuous movements and feet penetrating through the floor or sliding when on the ground.

A limitation of these works is that their models are based on the Skinned Multi-Person Linear (SMPL) model, which does not use a biomechanically accurate or interpretable kinematic chain \citep{loper_smpl:_2015, keller_skin_2023}. Furthermore, these methods have not been tested on a dataset of people with movement impairments, nor have their forces been validated. \citet{smyrnakis_advancing_2024} addressed this later concern by testing video-driven imitation learning applied to a large dataset of gait from a motion analysis laboratory, and found that incorporating physics into the inference process improved the accuracy of estimated spatiotemporal gait parameters over model-free inference. However, this work used a simplified model for humans rather than a biomechanical model, limiting the tracking accuracy. Concurrent work to ours showed that imitation learning could be applied to a muscle-driven lower body biomechanical model trained on a dataset of able-bodied movements.

\section{Approach}

Our system aims to closely replicate kinematic trajectories in a physics simulation of a human biomechanical model. The trajectories consist of a sequence of discretely sampled poses and pose velocities: $\mathbf \tau = \left\{ \boldsymbol \beta, \mathbf  q_0, \dot {\mathbf  q}_0, \mathbf q_1, \dot  {\mathbf  q}_1 ... \mathbf q_T, \dot  {\mathbf  q}_{T} \right\}$, which are obtained from markerless motion capture of human movement and $\boldsymbol \beta$ are parameters that represent the anthropomorphic measurements of the participant.

\subsection{Reinforcement learning preliminaries}

Our goal-conditioned policy takes the current state of the biomechanical model in the physics simulation, $\mathbf s$, and future samples observations from the trajectory, $\mathbf o$, and outputs an action $~\mathbf a \sim \pi(\mathbf s, \mathbf o)$. This action is passed to the physics simulator, which produces the next state, $\mathbf s_{t+1} = \mathcal T (\mathbf s_t, \mathbf a_t)$. The policy receives a reward based on the state, action, and the trajectory with a greater reward for more closely following the trajectory: $r(\mathbf s_{t+1}, \mathbf a_t, \boldsymbol \tau)$, which we describe in more detail below. The policy is trained to maximize the expected reward with a discount factor $\gamma$:
\begin{equation}
\mathcal J(\pi_\theta) = \mathbb E_{\boldsymbol \tau \sim \mathcal D, \, a \sim \pi_\theta(\mathbf s, \mathbf o)} \left[  \sum_{t=0}^T \gamma^t r(\mathbf s_t, \mathbf a_t, \boldsymbol \tau) \right].
\end{equation}

\subsection{Biomechanical model}

Our biomechanical model is based on LocoMujoco \citep{al-hafez_locomujoco_2023}. We used both the torque-driven model and a muscle-driven model with 92 muscles in the lower limbs. An essential step for accurately imitating movement sequences of real individuals is to match their individualized anthropomorphic measurements during the physics rollout.  Our imitation learning environment applies a scaling parameter to the biomechanical model corresponding to the specific trial being replicated.

\subsection{Residual force control}

Many of the trajectories in our dataset cannot be replicated exactly in a physics simulator due to forces external to the biomechanics, such as rolling walkers, canes, forces applied by therapists, or contact with chairs. Without any additional changes, the imitation learning policy would best replicate these movements by introducing significant kinematic deviations to make them physically plausible without additional forces. This contrasts with our goal of close kinematic tracking. Residual force control (RFC) enhances tracking despite such a sim-to-real gap with a non-physical 6-degree-of-freedom force to the pelvis \citep{yuan_residual_2020}. These additional forces are included in the action space of the model.

\subsection{Biomechanical model observation vector}

The policy receives both the state of the model and the future observations. The state of the model is parameterized identically to the base Brax humanoid environment that our work is based on. Specifically, it is the \texttt{qpos} vector reflecting the pose of the model after excluding the horizontal components and the \texttt{qvel} vector that includes the derivatives of the \texttt{qpos}, $(\hat {\mathbf q}_t,\hat {\dot {\mathbf q}}_t)$. We use the hat to indicate the current simulator state versus the trajectory to be replicated. The \texttt{qvel} vector is one element shorter than \texttt{qpos} as the quaternion orientation becomes a 3-element derivative. The next element is the mass and inertia of the 21 body elements relative to the center of mass, each of which has 10 elements, followed by the center of mass-based velocities for those bodies, each of which has 6 elements. Finally, it includes the active forces across each degree of the 40 degrees of freedom.

The next set of variables passed to the policy corresponds to observations about the trajectory.\newline
\begin{equation}
\mathbf o_t = \left\{ \mathbf a_{t-1}, \beta, \mathbf q_{t}, (\mathbf q_{t+k} \ominus \hat {\mathbf q}_t, \dot {\mathbf  q}_{t+k} \ominus \hat {\dot {\mathbf q}}_t) \; \forall \; k \in K \right\}
\end{equation}
We typically used $K=\left\{ 0,1,2,4,8,16 \right\}$, allowing the policy to plan several timesteps into the future. Each of these observations is represented as the difference between the future target location and the current simulation state, and we use the $\ominus$ operator to indicate that the orientation component of the pelvis (root) pose computes the quaternion difference between the current and target orientation. Finally, the observation vector includes the body scale parameter for that trajectory, $\beta$, and the actions vector from the prior time step, $\mathbf a_{t -1}$.

\subsection{Reward model}

Our reward function, $r(\mathbf s_{t+1}, \mathbf a_t, \boldsymbol \tau)$, has several components, most formulated as negative rewards for tracking errors. The first two losses penalize deviations from the next state and the target next state angles and velocities.
\begin{equation}
l_{\mathbf q} = (\mathbf q_{t+1} \ominus \hat {\mathbf q}_{t+1})^2  \cdot \, w_{\mathbf q}
\end{equation}
\begin{equation}
l_{\dot{\mathbf q}} = (\dot{\mathbf q_{t+1}} \ominus \hat {\dot {\mathbf q}}_{t+1})^2  \cdot \, w_{\dot {\mathbf q}}
\end{equation}
The $w_{\mathbf q}$ and $w_{\dot{\mathbf q}}$ vectors allow us to apply different weights to different components of the loss, such as to the root state elements versus the body state elements. For example, we do not penalize the absolute height of the pelvis to allow for slight differences between the interactions of the foot with the ground plane from the physics versus the room calibration.

We also include a reward to shape the actions. This includes penalizing having excessive actions, excessive changes in actions over time, and additional penalization to discourage using the RFC actions.
\begin{equation}
l_{\mathbf a} = w_{\mathbf a} \,  \Vert {\mathbf a_t} \Vert^2 \, + \,w_{\Delta \mathbf a} \, \Vert {\mathbf a_t} + {\mathbf a_{t-1}} \Vert^2 - w_{\mathrm RFC} \, \Vert {\mathbf a_t^{(0:6)}} \Vert^2
\end{equation}
\begin{equation}
r(\mathbf s_{t+1}, \mathbf a_t, \boldsymbol \tau)= r_{\text {alive}} - l_{\mathbf q} - l_{\hat{\mathbf q}} - l_{\mathbf a}
\end{equation}
Where $r_{\text {alive}}$ is a constant to make the reward generally positive. This formulation is different than most imitation learning works, which pass reward components through exponential non-linearities to ensure the reward is always positive. However, other work has shown that mean squared error produces more precise tracking \citep{smyrnakis_advancing_2024}.

\subsection{Imitation kinematic trajectories dataset}

Our dataset contains biomechanical trajectories from a wide range of participants performing different activities and a variety of motor impairments of different etiologies. Many of these activities are commonly used clinical outcome assessments for mobility, including overground walking, the timed-up-and-go test, the L-test, the four-square-step-test, and the functional gait assessments, such as which also involve sitting in a chair. Many of these participants had gait impairments, such as from a lower-limb amputation and the use of a prosthetic, or from neurologic conditions such as a stroke. Many of these had forces applied to their body other than through their legs, such as due to the use of an assistive device like a cane, crutch or walker, as well as forces from a therapist. Participants performing these tasks were recorded with multiple cameras and processed with markerless motion capture algorithms to generate the kinematic trajectories (see supplementary materials for details). The trajectory fitting procedure produces the body shapes and kinematic trajectories,  $\mathbf \tau = \left\{ \boldsymbol \beta, \mathbf  q_0, \dot {\mathbf  q}_0, \mathbf q_1, \dot  {\mathbf  q}_1 ... \mathbf q_T, \dot  {\mathbf  q}_{T} \right\}$, sampled at 30 Hz. During training, these trajectories are interpolated to 90 Hz. In trials, participants walked along an instrumented walkway. This floor detected the timing and location of foot contact events and foot off events, which produces paired data for validating the physics inference of gait event timing. All data collection was approved by our Institutional Review Board.

This dataset was split into a 90/10 training and validation set at the level of participants. Thus, all numbers reported below are for participants never seen by the policy during training, unless otherwise specified. The training split was 32 hours of markerless motion capture from 426 participants from 6564 trials. The test split was 2.3 hours of motion capture from 41 participants spanning 401 trials.

\subsection{Implementation details}

We used a policy depth of 8 layers and a width of 256 layers. The value function had a depth of 8 layers and a width of 1024 layers. We trained our model with the PPO \citep{schulman_proximal_2017} implementation from Brax \citep{freeman_brax_2021}. Training was run on an A100 and used 4096 parallel environments, which ran for 4B simulation steps for the torque-driven models and 8B simulation steps for the muscle-driven models.

After interpolation, our 30 fps markerless motion capture data was fed at 90 fps into the policy. The simulator then took 5 simulation substeps per frame to run the physics at 450 Hz. Our supplementary materials describe our remaining hyperparameters.

\section{Experiments}

\subsection{Torque-driven model evaluation}

In general, our torque-driven model closely tracked the kinematics from our markerless motion capture system across a range of activities and severities of gait impairments. Figure~\ref{fig:168_torque} shows an example target and replicated trajectory, along with the joint velocities and corresponding internal joint moments and ground reaction forces.

We evaluated our models through several metrics. First, we measured the mean absolute error between the target trajectories and the physics-driven replication for each trial and report the average over trials in Table~\ref{table:metrics}. We also tracked the failure percentage as the number of trials that could not be completely replicated due to the pelvis height below 0.3 m. This excluded 4 trials from being tested, as participants were doing therapy assessments on the floor, thus the table scores falures out of the remaining 397 trials in our test set). In brief, our baseline policy tracked within 4 cm of the target trajectory and tracked joints with less than a degree of average error.

\begin{table}
\centering
\begin{tabular}{p{\dimexpr 0.111\linewidth-2\tabcolsep}p{\dimexpr 0.111\linewidth-2\tabcolsep}p{\dimexpr 0.111\linewidth-2\tabcolsep}p{\dimexpr 0.111\linewidth-2\tabcolsep}p{\dimexpr 0.111\linewidth-2\tabcolsep}p{\dimexpr 0.111\linewidth-2\tabcolsep}p{\dimexpr 0.111\linewidth-2\tabcolsep}p{\dimexpr 0.111\linewidth-2\tabcolsep}p{\dimexpr 0.111\linewidth-2\tabcolsep}}
\toprule
Model & Failure Rate (\%) & Pelvis Hor (mm) & Pelvis Hor Vel (mm/s) & Joint Angle (deg) & Joint Vel (deg/s) & Hip (deg) & Knee (deg) & Ankle (deg) \\
\hline
Baseline & 3.8 & 42 & 82 & 0.65 & 8.2 & 1.2 & 0.9 & 0.7 \\
-Future & 6 & 83 & 131 & 1.45 & 11.2 & 2.5 & 1.7 & 1.2 \\
-RFC & 61.2 & 430 & 287 & 1.49 & 15.5 & 3.1 & 1.9 & 2.2 \\
Muscle & 9.6 & 38 & 100 & 1.68 & 14.2 & 2.1 & 1.7 & 3.5 \\
\bottomrule
\end{tabular}
\caption{Tracking errors between target kinematic and imitation learning rollouts \label{table:metrics}}
\end{table}

\subsection{Instrument walkway comparison}

\begin{figure}[!htbp]
\centering
\includegraphics[width=0.75\linewidth]{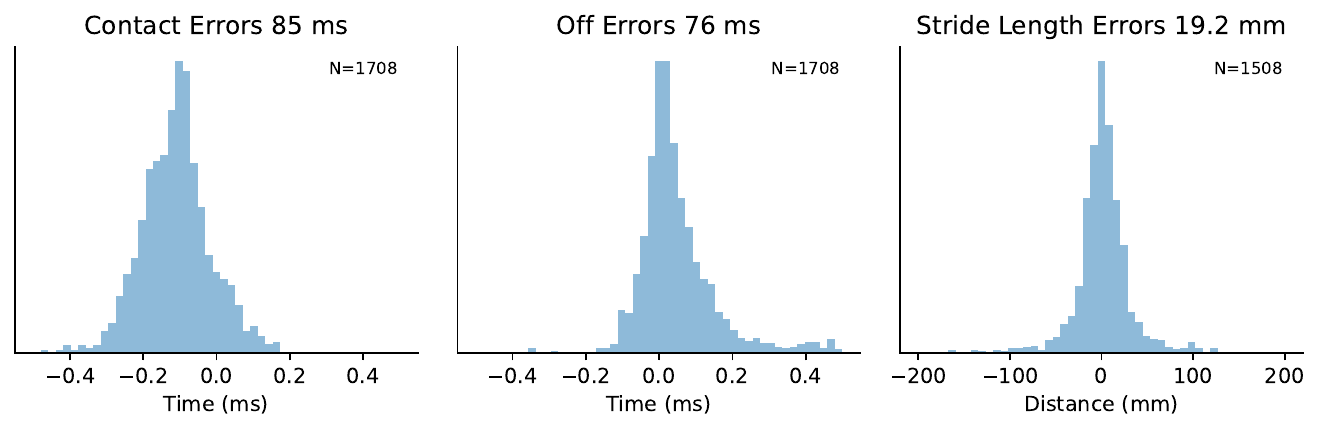}
\caption[]{Histogram of errors for foot contact events, foot off events, and stride lengths, compared to an instrumented walkway.}
\label{fig:gaitrite_errors}
\end{figure}

79 of these trials involved walking over an instrumented walkway that detects foot contact and lift-off timing and location, and we measured the error between these ground truth events and our vertical ground reaction forces. Figure~\ref{fig:gaitrite_errors} shows the histogram of these errors. We quantified this with the median error and normalized interquartile range (NIQR; 0.7413 times the interquartile range). We also computed the center of pressure from the ground reaction forces, which is the weighted sum of the location of any foot contact geometries with the floor, weighted by the vertical component of the ground reaction force. For each gait cycle, we computed the difference between the stride lengths (forward progression across the instrumented walkway from one side of the body making contact with the ground and the next) from the center of pressures and compared them to the stride lengths computed by the instrumented walkway. Figure~\ref{fig:gaitrite_errors} shows the histogram of errors, which we also summarized with their medians and NIQRs. These errors are reported for all our models in Table~\ref{table:gaitrite_metrics}.

\begin{table}
\centering
\begin{tabular}{p{\dimexpr 0.250\linewidth-2\tabcolsep}p{\dimexpr 0.250\linewidth-2\tabcolsep}p{\dimexpr 0.250\linewidth-2\tabcolsep}p{\dimexpr 0.250\linewidth-2\tabcolsep}}
\toprule
Model & Foot Contact (ms) & Foot Off (ms) & Stride (mm) \\
\hline
Baseline & 80 & 71 & 20 \\
-Future & 101 & 53 & 27 \\
-RFC & 185 & 41 & 179 \\
Muscle & 120 & 135 & 29 \\
\bottomrule
\end{tabular}
\caption{Errors compared to an instrumented walkway from KinTwin \label{table:gaitrite_metrics}}
\end{table}

\subsection{Failure cases}

We had 15 out of 397 trials where our baseline torque-driven imitation learning policy could not replicate complete trajectories. Three were a pediatric patient walking with a rolling walker in a very leaned-forward way, where the model tripped and then stumbled (see supplementary materials Figure~\ref{fig:pediatric_trip}). Several others corresponded to a manual wheelchair user, where the RFC did not reliably allow them to float. The remaining corresponded to motion capture errors with implausible and impossible movements.

\subsection{Lesions and hyperparameter testing}

We also performed two lesion studies reported in Table~\ref{table:metrics} and Table~\ref{table:gaitrite_metrics}. First, we disabled future observations of the trajectory. Unsurprisingly, this worsened the tracking performance as the policy does not have the information in order to plan ahead. It also increased the failure rate. In the second study, we disabled RFC. This had an even more pronounced worsening of both of these metrics. This is unsurprising because many of our trials cannot be completed without additional forces, such as sitting and rising from a chair. The instrumented walkway errors also worsened, where people were always standing, but still may be using additional forces, such as from a therapist or assistive device. We do note that this policy was continuing to steadily improve at 4B steps, and we did not attempt to further optimize this training.

\subsection{Muscle-driven model evaluation}

Next, we repeated this experiment with our muscle-driven model. As with our torque-driven model, our muscle-driven model was able to replicate our kinematic trajectories, although not as accurately as our torque-driven model. Table~\ref{table:metrics} shows an increase in the failure rate and the joint angle tracking error, increasing from 0.65 degrees to 1.68 degrees. The vertical GRF timing error also increased to 120-135ms and the stride length error increased from 20 to 29mm (Table~\ref{table:gaitrite_metrics}).

\begin{figure}[!htbp]
\centering
\includegraphics[width=0.8\linewidth]{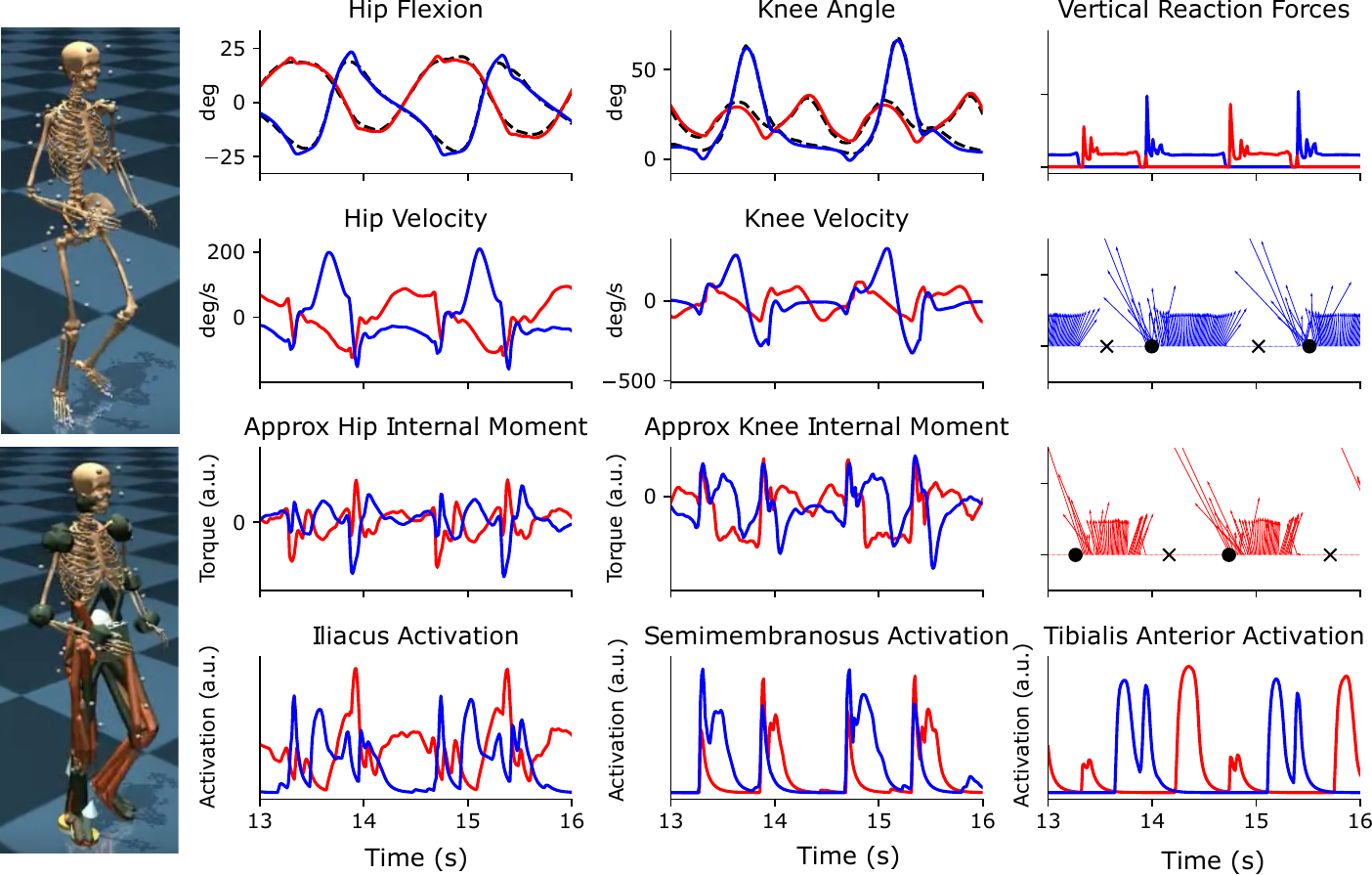}
\caption[]{Example muscle-driven model replicating walking from someone with a right (red traces) hemiparetic gait after a stroke. The top panels for the hip and knee show the target trajectory with a dashed black line, showing that the rollout closely replicates the target kinematics with their asymmetric pattern. The bottom traces show the muscle activations. For example, it is apparent that the semimembranosus (hamstring) shows reduced activation on the right side, corresponding to the reduced knee flexion on this side. Similarly, the tibialis anterior activation is inferred when initiating the swing phase to clear the toes on and during weight acceptance to prevent foot slap while capturing the asymmetric dynamics and magnitude.}
\label{fig:13413_muscle}
\end{figure}

Figure~\ref{fig:13413_muscle} shows an example rollout along with the corresponding muscle activations, which shows the imitation learning policy infers both plausible timings of the muscle activations as well as capturing asymmetrical magnitudes and timings that are consistent with this asymmetric, hemiplegic gait pattern after a stroke.

\subsection{Models are sensitive to clinically meaningful differences}

While there is a great deal of work to do for further validating the inferences from our imitation learning policies, our initial studies already show this system is sensitive to clinically meaningful features. For example, Figure~\ref{fig:168_torque}  shows a participant with a left transfemoral amputation using a prosthesis. The waveforms from the left (blue) show reduced extensor torques at the hip and reduced forward propulsion in the ground reaction force at terminal stance, while also detecting the brief extensor torque causing a negative hip flexion velocity that ensures the prosthetic knee fully extends prior to initial contact. It also infers smaller torques generated in the prosthetic knee compared to the intact knee Similarly, Figure~\ref{fig:13413_muscle} shows the muscle-driven policy detects asymmetrical muscle activations from an individual neurologic gait impairment.

\section{Discussion}

Here we show that imitation learning of human movement can be translated onto torque-driven and muscle-driven biomechanical models. We found that both classes of models could accurately track the kinematics of both able-bodied movement and those from individuals with gait impairments, although accuracy was higher with the torque-driven model.

In contrast to prior work on imitation learning, we focused on the use of biomechanical models and on training a single policy that would precisely track movements for a wide range of mobility assessments and people with varying motor impairments. Most works on imitation learning for human movement train on movement datasets that only include able-bodied individuals and use body models that are derived from the SMPL family \citep{loper_smpl:_2015}, which do not have an anatomically accurate kinematic model \citep{keller_skin_2023}. One exception is \citet{lee_scalable_2019}, which used reinforcement learning to train a policy for replicating a single kinematic trajectory using a full-body muscle-driven model with 346 degrees of freedom.

Concurrent work to ours used imitation learning of a muscle-driven biomechanical model of the lower body from MyoSuite \citep{caggiano_myosuite_2022, simos_reinforcement_2025}.
Our work differs in several notable ways. Firstly, they used a dataset that only contains data from able-bodied individuals \citep{simos_reinforcement_2025}. Additionally, they did not account for body scaling, nor does their model does not track the movement of the arms. They also did not cross validate their policy or report joint angle tracking accuracy or the accuracy of gait event timing, making direct comparison challenging. However, their reported mean joint position errors (45mm) was higher than both our mean pelvis position horizontal error (42 mm) and our stride length errors (20mm). They also included a number of additional features that we did not find necessary including 1) injecting noise into the penultimate layer of the policy to produce more structured exploration, 2) including a proportional-derivative controller between the policy and the model rather than allowing the policy to control the muscles directly, 3) using a mixture of experts in their controller, and 4) finally using hard-negative mining during training. We did use RFC due to the external forces in our dataset, which they did not include, and this does make the learning process substantially easier. Additionally, we used the MuJoCo MJX GPU acceleration of the imitation learning for both muscle and torque driven models, which allowed us to train the muscle-driven model on an A100 GPU with 8B simulation steps in 2 days in contrast to 1B simulation steps over 10 days using 128 CPUs for the simulations and an A100 for the policy updates.

In addition to the accurate tracking of kinematics, the timing and location of the inferred vertical ground reaction forces show close agreement with an instrumented walkway across a population of individuals with a wide range of gait impairments. While there is much more work to do to tune and validate the torques and muscle activations, our initial studies show these produce plausible values that are sensitive to clinically meaningful features of the participants. For example, detecting asymmetries of muscle activations in individuals with neurologic conditions or detecting the reduced propulsion from the intact side of a transfemoral amputee. In future work, we anticipate both comparing these results against formal motion capture that measures ground reaction forces and uses this to compute joint moments. We also anticipate systematically evaluating the differences between participants and mapping this onto their clinical context.

We see this work as the first step towards creating digital embodiments or digital twins that explain why people move a certain way, and begin to scratch the surface below kinematics by inferring torques and muscle activations. This opens many promising directions, such as distilling these policies into generative models \citep{luo_universal_2024}, or building more neuro-mechanistic models that incorporate spinal reflex circuits and higher-level neural control of movement \citep{lassmann_dysfunctional_2023}.

\subsection{Limitations}

However, our tracking results are not perfect. Some problems arise from errors in the markerless motion capture we use, which has been shown to have about 10-15mm of error at room-scale tracking \citep{cotton_differentiable_2025}. These errors may force the policy to track a trajectory that is not physically plausible, such as the toe penetrating the ground during the swing phase, resulting in a trip during policy rollout. Other errors may require improvements in our training strategy, model, or more data. One benefit of the MuJoCo physics simulator underlying our work is its differentiable nature. This can be exploited by differentiation-aware reinforcement learning policies like short-horizon actor critic, which has specifically been shown to improve the accuracy of trajectory following \citep{xu_accelerated_2022, georgiev_adaptive_2024}. Alternatively, these errors may also arise if the muscle-driven model lacks the strength to support the pelvis with a more medial foot placement, such as due to weak hip abductors. We defer optimization of the musculoskeletal model and use these differentiable learning policies for future work.

While tracking the kinematics worked quite well, we anticipate substantial work will be required to fully tune both the torques and muscle activations. For example, scaling the masses and moments of inertia of body segments is standard in marker-based motion capture and could be incorporated into our approach in a straightforward manner. Additionally, we saw some muscles hitting their maximum activation during regular walking.  We also only evaluated the accuracy of the thresholded vertical component of the GRF. However, the simulation produces a 3-dimensional GRF that includes the forward component that propels movement. Qualitatively, our waveforms demonstrated the traditional ``butterfly pattern'', but truly validating these 3D GRFs against force plate measurements is an important future direction. This temporal resolution is generally sufficient for dividing the gait into cycles, but is likely insufficient for clinically meaningful differences in the percentage of time participants might spend in different phases of the gait cycle. We suspect optimizing the foot contacts, such as to account for the variety of shoes people wear, will be an important future direction. Possibly related to this, for some versions of our muscle-driven models, there was often a brief bounce at initial contact. Our dataset contains a subset of participants with wearable electromyography sensors, which could allow incorporating muscle activations into the reward function and improving the accuracy of these. We leave this further tuning for future work.

We also suspect there are opportunities to improve how motion trajectories are represented and passed to the policy. In our current work, all of the future observations are concatenated and passed into an MLP. Prior work has shown the benefit of learning a motion representation for imitation learning, and can even produce a generative model of movement \citep{tessler_calm_2023, luo_universal_2024}.

\subsection{Potential broader impact}

From our perspective of clinical access to motion analysis, we anticipate these methods primarily producing societal benefits. Furthermore, our focus on validation for a population that includes able-bodied and individuals with movement impairments is an important step towards AI Fairness for people with disabilities. Nonetheless, all technologies have the potential for misuse. One concern is the premature use of this technology for clinical decision making, and we strongly emphasize that these tools can be challenging to use in the wild, and caution is required at all stages. Furthermore, there are many movement patterns and diagnoses not represented in our current dataset. Another concern is that this technology could allow inferring private information about health status from video observations without consent.

\section{Conclusion}

Imitation learning using neuromuscular models can closely track the kinematics of human movement from participants with a range of motor impairments using both torque-driven and muscle-driven models. The policy learns to solve the inverse dynamics problem of inferring the mechanisms underlying movement, and is sensitive to clinically meaningful differences. This offers a very promising approach for creating digital twins based on clinically accessible motion capture and better phenotyping the mechanics of human movement.

\begin{ack}
This work was supported by R01HD114776 (RJC) and the Research Accelerator Program of the Shirley Ryan AbilityLab (RJC).
\end{ack}

\bibliographystyle{unsrtnat}
\bibliography{main.bib}

\section{Technical Appendices and Supplementary Material}

\subsection{Biomechanical model}

We modified the LocoMujoco model, which was derived from \citet{hamner_muscle_2010}, to only keep the collisions on the feet and added a neck joint with 3 degrees of freedom. We replaced the Euler parameterization of the pelvis with a quaternion. The joint limits were increased to match those used in our markerless motion capture fits.  LocoMujoco has both a torque driven model and a muscle driven model. Both have 21 body elements (excluding the floor) and 40 degrees of freedom (dof), with the pelvis location and orientation having 7 parameters to represent 6 dof with the quaternion rotation, and 34 dof being joints across the body. The torque driven model includes 40 actuators, with 6 of those being the residual force control (RFC) \citep{yuan_residual_2020} applied to the pelvis and the remaining 34 producing torques at each joint. The muscle driven model has a total of 118 actuators with 6 being RFC, 20 being torque actuators above the pelvis, and 92 being muscles in the lower body.

Our markerless motion capture dataset includes 8 body scale parameters for each individual: an overall size, the pelvis, left thigh, left leg and foot, right thigh, right leg and foot, the left arm, and the left leg, which were also applied to the biomechanical model in the imitation learning environment.  For the muscle-driven model, we also rescaled the muscle lengths proportionately to keep them functioning similarly. This was done by adjusting the \texttt{tendon\_length0} parameter for each muscle to the length determined when the scaled model was set to the \texttt{qpos0} (neural) pose position.

\subsection{Biomechanical Reconstruction}

We performed an end-to-end optimization of kinematic trajectories with respect to detected keypoints, similarly to the approach described in \citep{cotton_differentiable_2025}.  Briefly, the end-to-end optimization process leverages a GPU-accelerated, fully differentiable biomechanical model, implemented in MuJoCo, to fit kinematic trajectories directly from markerless motion capture data. This method uses an implicit representation (a multi-layer perceptron) that maps time to pose parameters ($f_{\phi}(t) \rightarrow \theta_t$). These pose parameters are then fed into the forward kinematic model, $\mathcal M(\theta,\beta)$, that also incorporates individualized body scaling ($\beta)$ and marker offsets. The core of the optimization is minimizing the reprojection loss, which is the error between the 2D keypoints detected in the camera images and the 3D virtual marker locations from the biomechanical model after they are reprojected back into each camera's 2D image plane. The keypoints used are the MOVI keypoints outputs from the MetrABs-ACAE algorithm \citep{ghorbani_movi_2021, sarandi_learning_2023}. This differentiable framework enables bilevel optimization, where skeleton scaling, marker offsets, and joint angle trajectories are jointly solved across multiple movement sequences for an individual participant.

One key difference between our approach and the one described in \citep{cotton_differentiable_2025} is that we also fit the derivative of the implicit representation with respect to time to obtain velocities, which we used in our imitation learning environment. This was optimized by taking a single integration step forward in time and adding a regularization to ensure the next pose sample aligned with the estimated one at that time step.

Trials were discarded if there the reprojection error from the reconstruction exceeded 20 pixels. The biomechanical model used to reconstruct their kinematic trajectories was the same one used in the imitation learning environment.

\subsection{Trajectory Augmentation}

We performed two types of augmentation to our data. The first was how we sampled trials. Our maximum rollout length was 600 steps (200 frames before interpolation). We randomly sample the start point of transitions from any time point between the beginning of the trial, followed by centering the trajectory at a random point within that trial. Finally we apply a random rotation and translation to the rollout.

The second type of augmentation we perform is adding noise to the initial joint \texttt{qpos} and \texttt{qvel}, to encourage the policy to track trajectories closely even after deviating. The noise is sampled from uniform distributions with widths $\sigma_{\mathbf q_0}$ and $\sigma_{\dot{\mathbf q}_0}$, respectively.

\subsection{Instrumented Walkway Errors}

We measure the error of the ground reaction forces from the physics simulation to the ground truth gait events. An affine spatial transformation and the timing offset between the markerless motion capture coordinates was used to align the modalities.

We identified when the GRF went above or below a threshold to identify foot contact and lift-off events in our physics simulations. For each foot contact and lift-off event from the instrumented walkway, we then identified the closest corresponding transition in the GRF and measured the temporal error.

We determined the center of pressure as the the weighted sum of the location of any foot contact geometries with the floor, weighted by the vertical component of the ground reaction force. For each gait cycle, we computed the different between the stride lengths (forward progression across the instrumented walkway from one side of the body making contact to the ground and the next) from the center of pressures and compared them to the stride lengths computed the by the instrumented walkway.

\subsection{Hyperparameters}

The simulation environment was configured with specific reward weightings to guide the learning process. The $r_\text{alive}$ was set to 10.0 and trials terminated if the pelvis height went outside the range (0.3, 2.0). The weight for most elements of the the pose imitation loss, $w_{\mathbf q}$, was 10.0, applied to the squared error between the target and current joint positions. The contribution of the pelvis's vertical position error to this loss was effectively removed by setting its specific weight, to 0.0. The error in pelvis orientation quaternion was weighted by 2.0 within the $l_\mathbf q$ loss, with a quaternion difference computed between the target and current quaternion. or the velocity imitation loss, $l_{\dot{\mathbf q}}$, the error in pelvis velocity was  set to 0.1, and the joint velocity error was weighted by 0.01. For the action loss, $l_{\mathbf a}$, the weight for the action cost, $w_\mathbf a$ was set to 0.01. The weight for the change in action cost, $w_{\Delta \mathbf a}$, was set to 0.01. The weight for the residual force control (RFC) action cost, $w_{\mathrm RFC}$, was set to 0.075. During trajectory initialization, recentering was enabled with random sampling and rotational augmentation noise of $\pi$ radians. Initial pose and velocity reset noise scales were set to 0.0.

Training was conducted for 4 billion simulation steps using 4096 parallel environments and a batch size of 1024. The number of minibatches per PPO epoch was 8. The learning rate started at 0.0002 and decayed using a cosine schedule with a final learning rate factor of 0.1  over 50,000 updates. The PPO clipping epsilon was 0.1, and the reward discounting factor $\gamma$ was 0.95. The unroll length for PPO was 10.

\newpage
\subsection{More torque examples}

Here we show several examples of the averaged kinematics and kinetics from the lower body of participants walking using the baseline torque-driven imitation learning policy. We show several examples of non-impaired and impaired gait. Note that this averaging over the gait cycle does not make gait timing apparent, and point out that the velocity scales vary greatly across participants. These plots also frequently average over trials where participants walk at different speeds.

\begin{figure}[!htbp]
\centering
\includegraphics[width=1\linewidth]{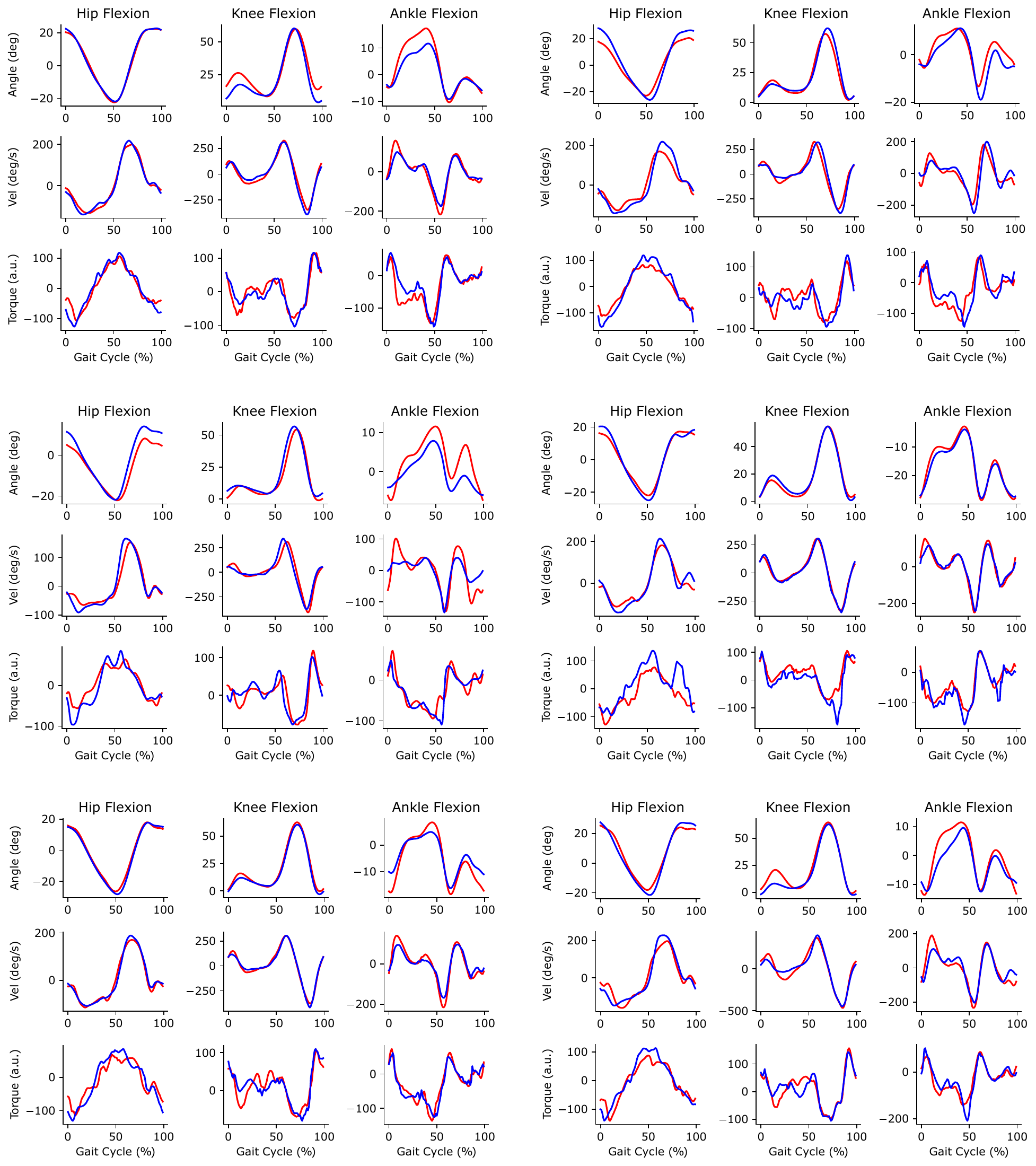}
\caption*{Six participants with no gait impairments. While some slight asymmetries are noted, likely due to personal walking patterns, these asymmetries are much smaller than the following participants with gait impairments.}
\end{figure}

\begin{figure}[!htbp]
\centering
\includegraphics[width=1\linewidth]{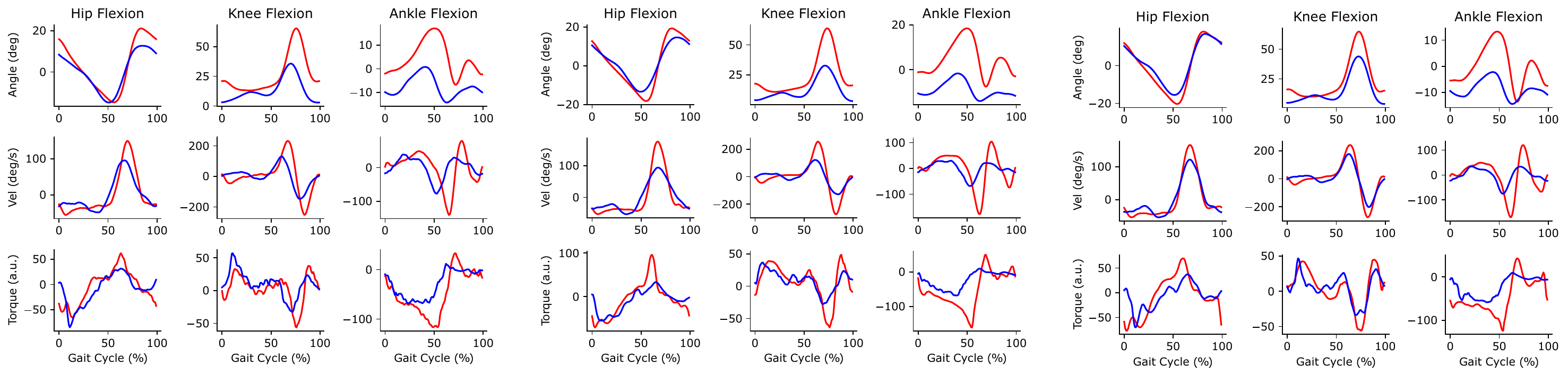}
\caption*{Participant with a left transfemoral amputation walking with a microprocessor knee prosthesis. The three panels show this participant on different days, consistently detecting a greater range of motion and more torque was detected on the intact (red) side.}
\end{figure}

\begin{figure}[!htbp]
\centering
\includegraphics[width=0.5\linewidth]{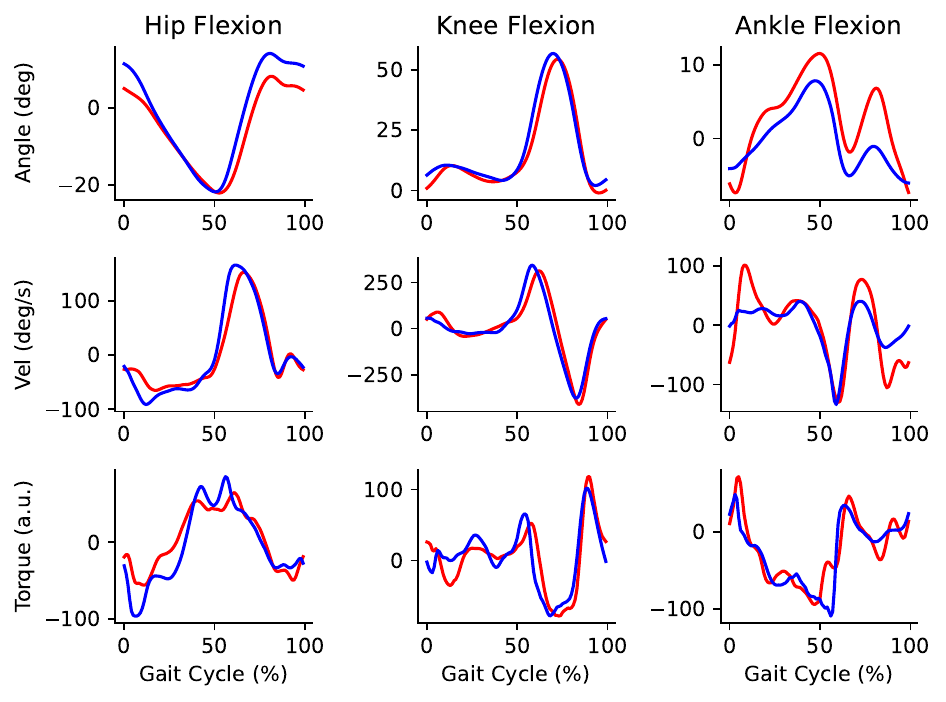}
\caption*{Participant with a left transtibial amputation. Compared to the transfemoral amputee, the hip and knee kinematics are more symmetrical, but asymmetry is notable at the prosthetic ankle..}
\end{figure}

\begin{figure}[!htbp]
\centering
\includegraphics[width=0.5\linewidth]{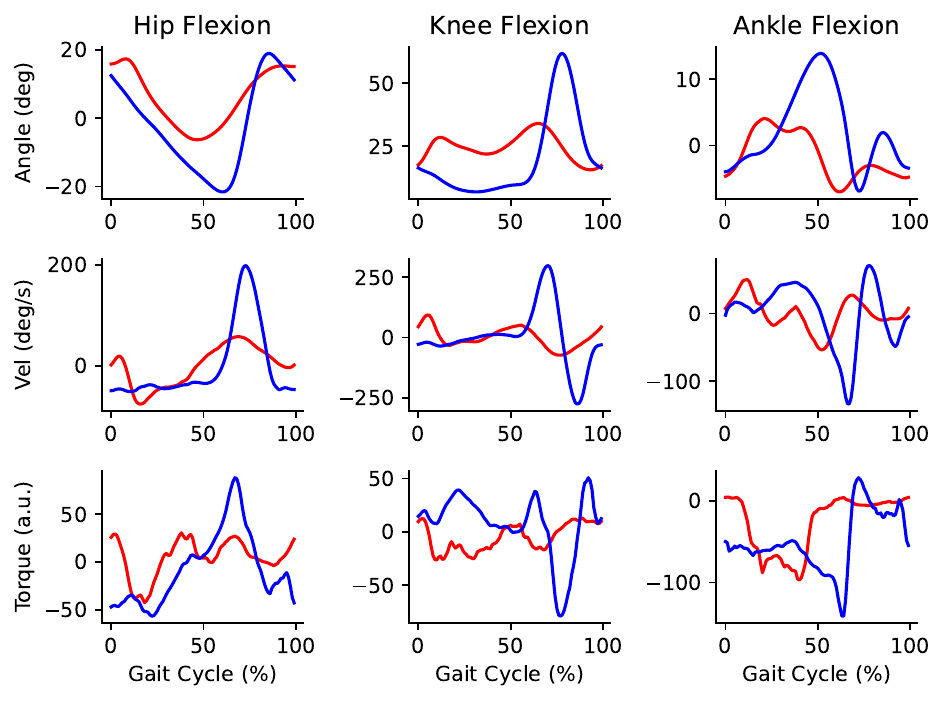}
\caption*{Participant with a stroke impacting their right side (red traces) resulting in a hemiparetic, stiff-knee gait pattern.}
\end{figure}

\begin{figure}[!htbp]
\centering
\includegraphics[width=0.5\linewidth]{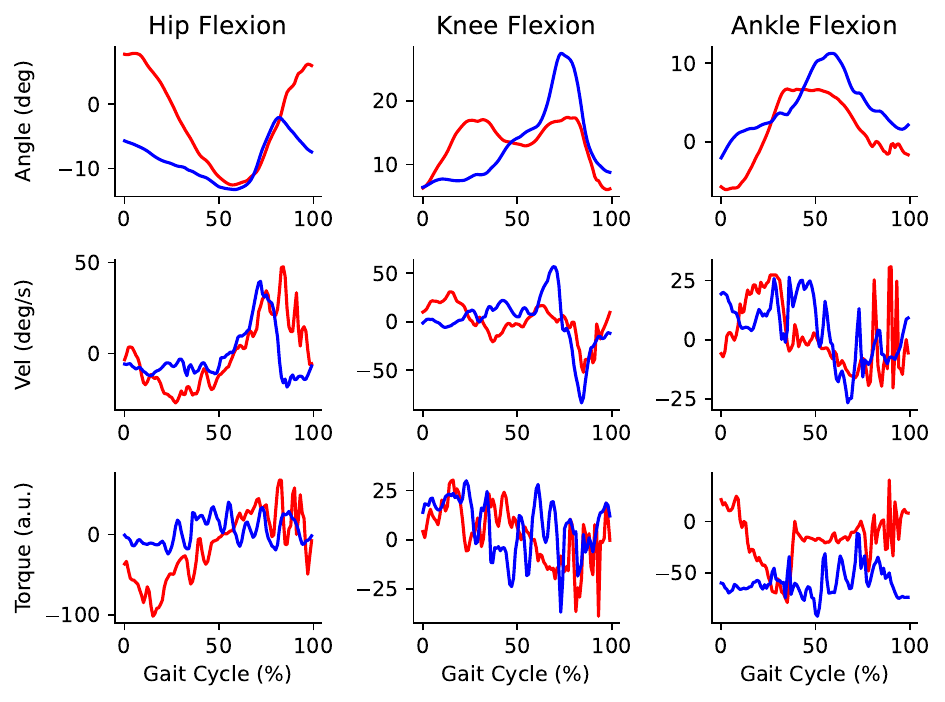}
\caption*{Participant with a spinal cord tumor.}
\end{figure}

\newpage
\subsection{More muscle examples}

Here we show several examples of the averaged kinematics, kinetics, and muscle activations.

\begin{figure}[!htbp]
\centering
\includegraphics[width=1\linewidth]{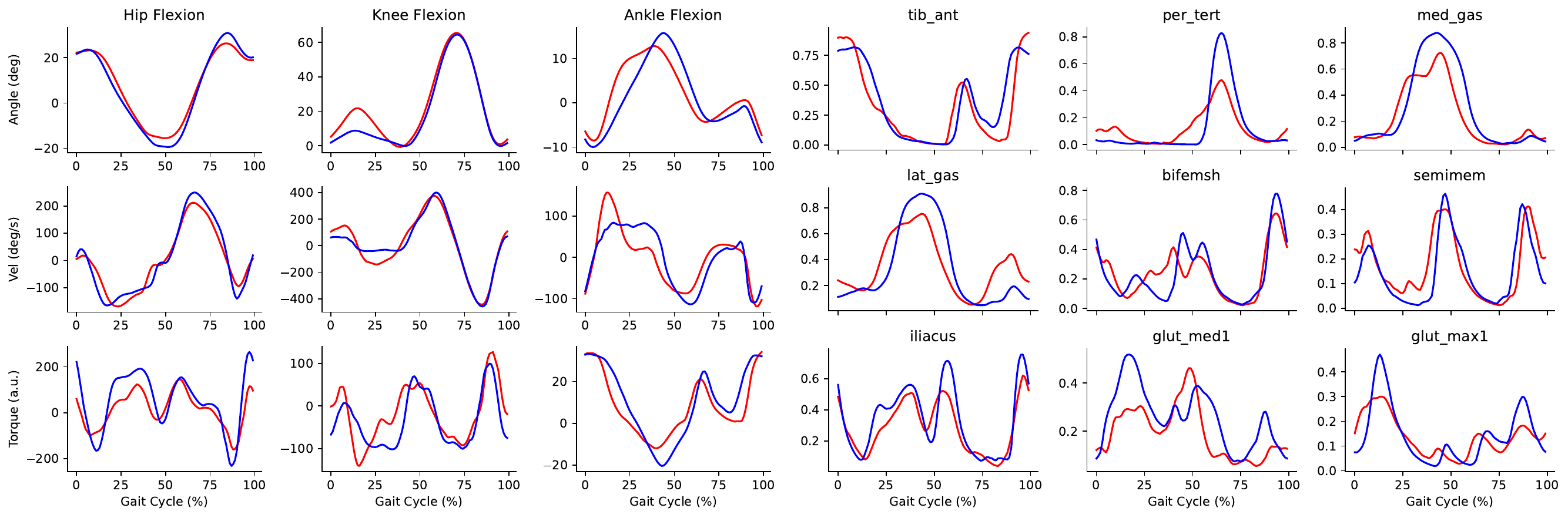}
\caption*{Participant with no gait impairments walking.}
\end{figure}

\begin{figure}[!htbp]
\centering
\includegraphics[width=1\linewidth]{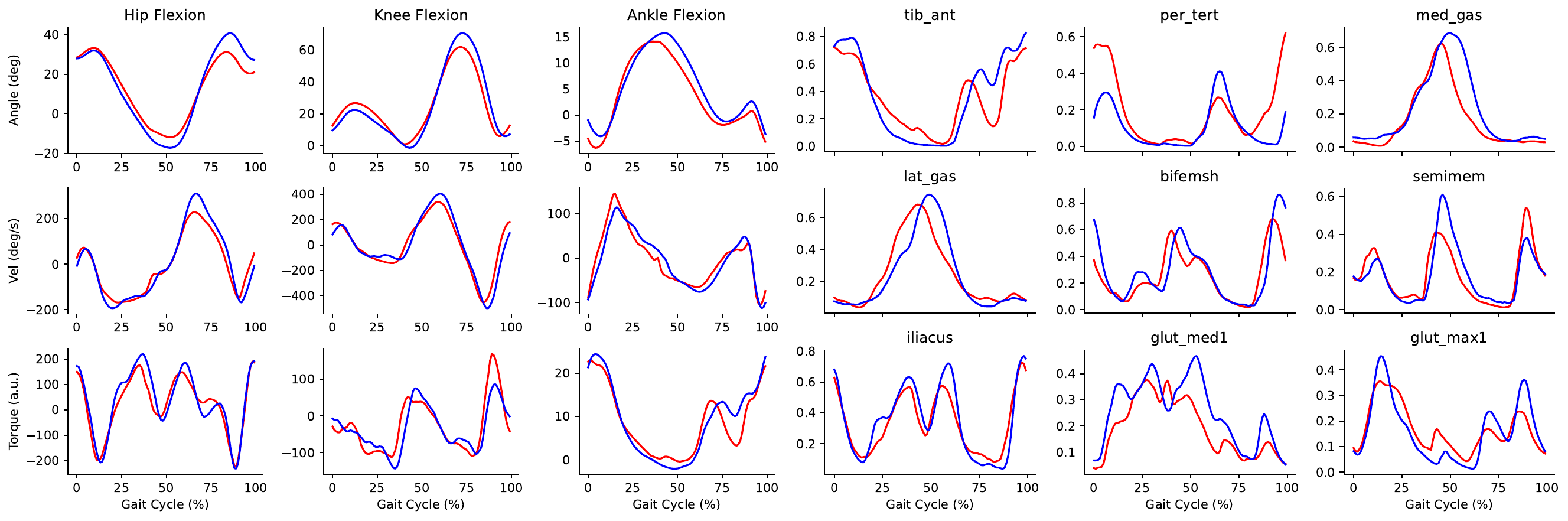}
\caption*{Participant with no gait impairments walking.}
\end{figure}

\begin{figure}[!htbp]
\centering
\includegraphics[width=1\linewidth]{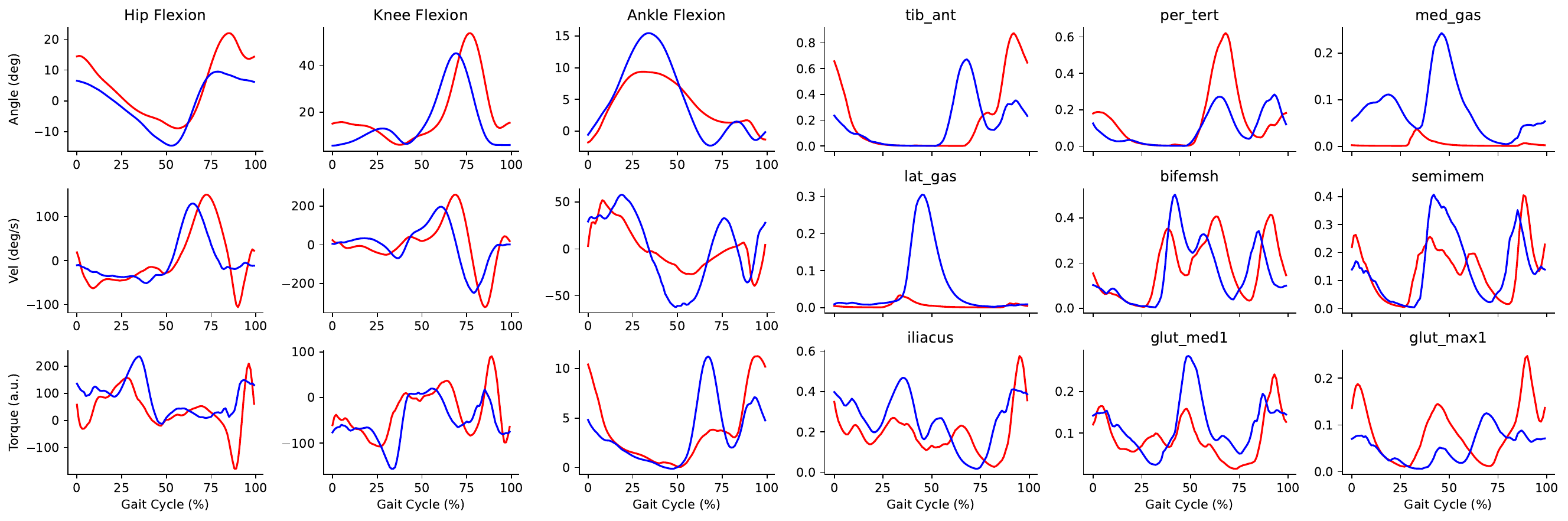}
\caption*{Participant with a history of stroke walking.}
\end{figure}

\begin{figure}[!htbp]
\centering
\includegraphics[width=1\linewidth]{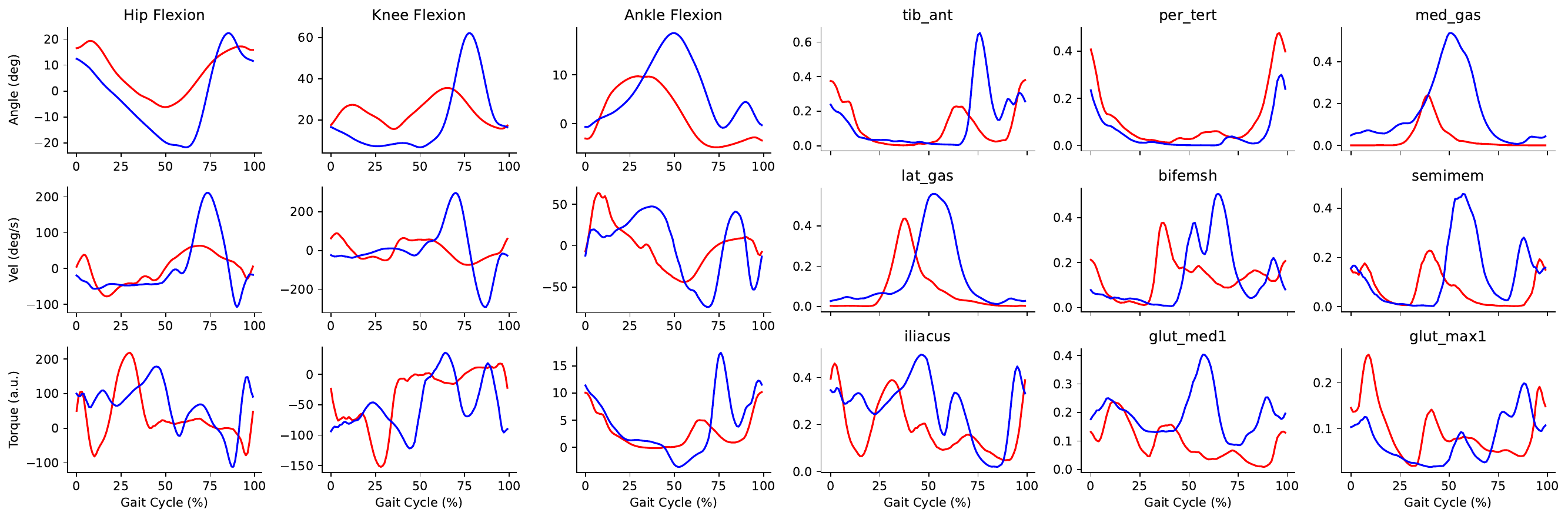}
\caption*{Participant with a history of stroke and right hemiparesis walking.}
\end{figure}

\newpage
\subsection{Failure case}

\begin{figure}[!ht]
\centering
\includegraphics[width=0.5\linewidth]{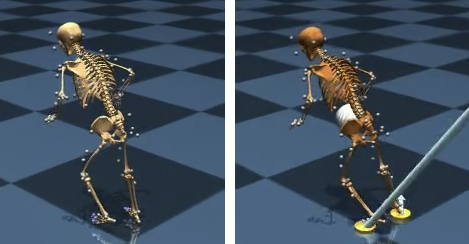}
\caption[]{Example of a failure for the imitation learning policy to replicate a pediatric walking sample using a rolling walker. Rendering from the target trajectory is shown on the left, and the policy tripping is shown on the right. Note the large posteriorly directed ground reaction forces as the foot catches the ground and initates a trip.}
\label{fig:pediatric_trip}
\end{figure}

\end{document}